\documentclass{article}
\usepackage[final]{neurips_2019}

\usepackage{times}
\usepackage{soul}
\usepackage{url}

\usepackage[hidelinks]{hyperref}
\usepackage[utf8]{inputenc}
\usepackage[small]{caption}
\usepackage{graphicx}
\usepackage{amsmath}
\usepackage{booktabs}
\usepackage{algorithm}
\usepackage{algorithmic}
\usepackage{wrapfig}
\usepackage{amsfonts}
\usepackage{multicol}
\usepackage{multirow}
\usepackage{amssymb}% http://ctan.org/pkg/amssymb
\usepackage{pifont}% http://ctan.org/pkg/pifont
\usepackage{xcolor}
\newcommand{\markc}{\color{green}\ding{51}}%
\newcommand{\markx}{\color{red}\ding{55}}%
\usepackage[colorinlistoftodos,textwidth=4cm,shadow]{todonotes}

\title{Sample Efficient Ensemble Learning with Catalyst.RL}
\author{%
  Sergey Kolesnikov\thanks{Moscow Institute of Physics and Technology} \\
  \texttt{scitator@gmail.com} \\
  Valentin Khrulkov\thanks{Skolkovo Institute of Science and Technology, Yandex}\\
  \texttt{khrulkov.v@gmail.com}
}

\begin{document}

\maketitle

\begin{abstract}

We present \texttt{Catalyst.RL}, an open-source \texttt{PyTorch} framework for reproducible and sample efficient reinforcement learning (RL) research. Main features of \texttt{Catalyst.RL} include large-scale asynchronous distributed training, efficient implementations of various RL algorithms and auxiliary tricks, such as $n$-step returns, value distributions, hyperbolic reinforcement learning, etc.
To demonstrate the effectiveness of \texttt{Catalyst.RL}, we applied it to a physics-based reinforcement learning challenge “NeurIPS 2019: Learn to Move - Walk Around” with the objective to build a locomotion controller for a human musculoskeletal model. The environment is computationally expensive, has a high-dimensional continuous action space and is stochastic. Our team took the 2nd place, capitalizing on the ability of \texttt{Catalyst.RL} to train high-quality and sample-efficient RL agents in only a few hours of training time. The implementation along with experiments is open-sourced so results can be reproduced and novel ideas tried out.

% In this paper, we present an approach to solve a physics-based reinforcement learning challenge “NeurIPS 2019: Learn to Move - Walk Around” with objective to build a controller for a musculoskeletal model with a goal of matching a given time-varying velocity vector field. The environment is computationally expensive, has a high-dimensional continuous action space and is stochastic. 
% To solve this issues we use \texttt{catalyst.RL}, an open source framework for RL research.
% Main features of \texttt{catalyst.RL} include large-scale asynchronous distributed training, efficient implementations of various RL algorithms and auxiliary tricks, such as frame stacking, $n$-step returns, value distributions, hyperbolic reinforcement learning, etc.
% Our team took the 2nd place, capitalizing on the ability of \texttt{catalyst.RL} to train high-quality and sample-efficient RL agents.

\end{abstract}

\section{Introduction}\label{sec:intro}
Within the last five years, a huge breakthrough has been made in deep reinforcement learning (RL). Autonomous agents, trained (almost) without any prior information about the particular games first matched and then surpassed professional human players in Atari $2600$~\citep{mnih2015human}, Chess~\citep{silver2017mastering2}, Go~\citep{silver2017mastering}, Dota~$2$~\citep{openaifive2018}, Starcraft II~\citep{deepmind2019}. Deep RL has also been successfully applied to neural networks architecture search~\citep{zoph2016neural} and neural machine translation~\citep{bahdanau2016actor}.

Despite these successes, all modern deep RL algorithms suffer from high variance and extreme sensitivity to hyperparameters~\citep{islam2017reproducibility,henderson2018deep} that makes it difficult to compare algorithms with each other, as well as to evaluate new algorithmic contributions consistently and impartially. For example, in ~\citep{fujimoto2018addressing} it has been shown that small changes of hyperparameters in DDPG~\citep{lillicrap2015continuous}, a popular baseline algorithm for continuous control, leads to a significant boost in its performance, which questions the validity of reported performance gain achieved over it in a number of subsequent works.

To make sure that the progress in deep RL research is evaluated consistently and robustly, we need a framework that combines best practices acquired within the last five years and allows a fair comparison between them. 

We present \texttt{catalyst.RL}, an open-source \texttt{PyTorch} framework for RL research written in \texttt{Python}, which offers a variety of tools for the development, evaluation, and fair comparison of deep RL algorithms together with high-quality implementations of modern deep RL algorithms for continuous control. Key features of our library include:
\begin{itemize}
    \item Distributed training with fast communication between various components of the learning pipeline and the support of large-scale setup (e.g., training on a cluster of multiple machines).
    \item Easy to use \texttt{yaml} configuration files with the complete list of hyperparameters for particular experiments such as neural network architectures, optimizer schedulers, exploration, and gradient clipping schemes.
    \item The ability to run multiple instances of off-policy learning algorithms simultaneously on a shared experience replay buffer to exclude differences caused by exploration and compare different algorithms (architectures, hyperparameters) apples-to-apples.
    %\item Efficient experience replay buffer implementation for off-policy methods with minimum necessary memory requirements and the support of frame stacking, $n$-step returns, and transitions prioritization.
\end{itemize}

To demonstrate the effectiveness of \texttt{catalyst.RL}, we applied it to the RL problem presented in \textbf{NeurIPS 2019: Learn to Move - Walk Around} competition. Participants were tasked with developing a controller to enable a physiologically-based human model to walk or run following velocity commands with minimum effort. Participants were provided with a human musculoskeletal model and a physics-based simulation environment (OpenSim \citep{delp2007opensim,seth2018opensim}) in which they synthesized physically and physiologically accurate motion. Entrants were scored based on how well the model moved according to the requested specified vector field of walking. 

Our method combines recent advances in off-policy deep reinforcement learning algorithms for continuous control, namely Twin Delayed Deep Deterministic Policy Gradient (TD3) \citep{fujimoto2018addressing}, quantile value distribution
approximation (QR-DQN) \citep{dabney2018distributional}, distributed training framework \citep{bellemare2017distributional, dabney2018distributional}, and parameter space noise with
LayerNorm for exploration \citep{ba2016layer}. Another important part of our solution was making use of recently introduced hyperbolic discounting \citep{fedus2019hyperbolic}.

The resulting algorithm scored a mean reward of 1346.939 in the final round and took 2nd place in NeurIPS'19 Learning to Move competition. 
Full source code\footnote{Full video with environment and agent demonstration is available at \url{https://youtu.be/WuqNdNBVzzI}} is available at \url{https://github.com/Scitator/run-skeleton-run-in-3d}.
% \todo{Our main contribution – novel usage of different DL (resnets) and RL techniques (td3, quantile value func approximation, hyperbolic gammas) for continuous-control action space environment with distributed training framework usage.}
%\input{text/2_related.tex}
%\input{text/challenges.tex}
\section{Framework}\label{sec:framework}

\begin{figure}[htb!]
\centering
\includegraphics[width=0.9\textwidth]{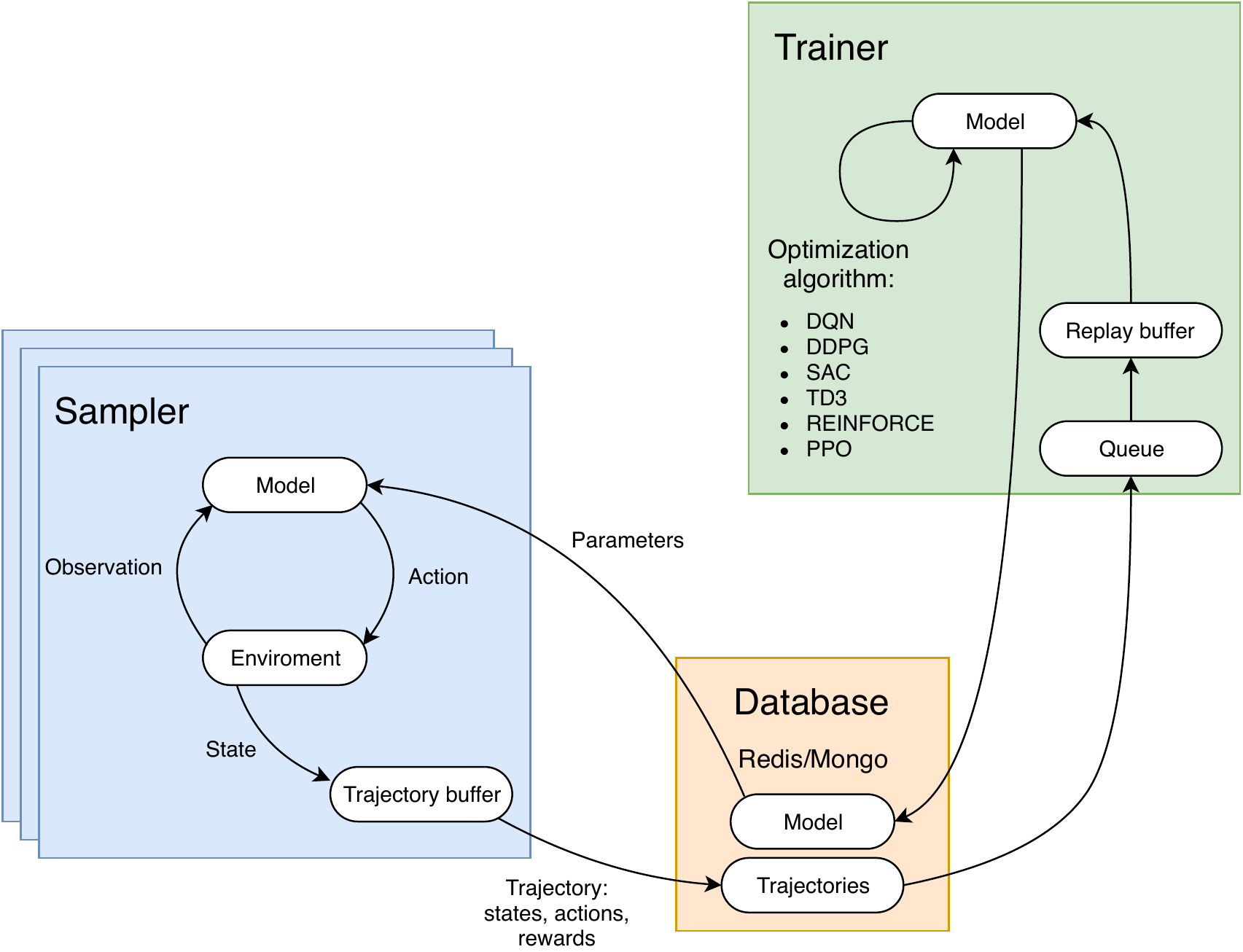}
\caption{\texttt{catalyst.RL} achitecture. Samplers interact with the environment and gather training data. Trainers retrieve collected data and update parameters of value function and policy approximators. All communication is conducted through a database.}
\label{fig:rl_arena}
\end{figure}

In this section, we introduce \texttt{catalyst.RL}\footnote{Find open-source code at \url{https://github.com/catalyst-team/catalyst}} and discuss the key design decisions which make it a suitable framework for flexible and reproducible deep RL research.

\paragraph{Flexibility} We believe that flexibility is a crucial aspect of any codebase suitable for RL research. The code should be readable and well-structured to minimize the efforts of the researcher who wants to adapt it to one's needs. 
\texttt{catalyst.RL} is developed in strict conformity with high-quality code standards that make it easy to change available algorithms or develop the new ones.

\paragraph{Scalability} Another important feature of our library is scalability. \texttt{catalyst.RL} can be easily used on a laptop, a single server or a huge cluster, which consists of multiple machines, without any code overhead capitalizing on its architectural design. More specifically, our solution is based on interconnected nodes of three types, namely \textit{trainers}, \textit{samplers}, and \textit{database}. Trainers are the nodes with GPUs which constantly update the parameters of neural networks representing policies and value functions. Samplers are CPU-based nodes that obtain policy networks weights and utilize them for inference only to collect training data. The database serves as an intermediary between trainers and samplers: it receives neural networks weights from trainers to send them to samplers; and receives data collected by samplers to send them for training. 
A critical feature of such a system is stability and robustness to individual breakdowns of its components, as all nodes are entirely independent and interchangeable.

\paragraph{Compositionality} \texttt{catalyst.RL} provides a convenient API for the development and evaluation of RL algorithms. In addition to the described above trainers, samplers, and database, which implement all necessary low-level operations for fast and efficient data transmission, it contains two additional abstract classes --- \textit{algorithm} and \textit{agent}. Algorithm implements the computation graphs of RL algorithms. Agent implements the neural network architectures used for policy (actor) and value function (critic) approximation.

Such compositionality allows for fast prototyping and easy debugging of RL pipelines, and usually, only a small part of the whole system needs to be changed. For example, to implement SAC with policies represented as normalizing flows~\citep{haarnoja2018latent}, we only need to add one new layer (coupling layer) and use it for the construction of a novel actor.

\paragraph{Reproducibility} To ensure reproducibility and interpretability of the results obtained by \texttt{catalyst.RL}, we predefine the same set of validation seeds for all models tested and compare them on this set only. To eliminate the evaluation bias coming from distorting actions produced by the policies with various exploration techniques, we run several samplers in deterministic mode and report their performance on the learning curves.

To keep track of various important metrics and visualize them during training we utilize the TensorboardX plugin. By default we log average reward, actor and critic losses, a number of data samples/parameter updates per second and any additional metrics of interest can be easily added by the user. We also provide a functional framework for saving the source code of each particular experiment, to exclude the situation when seemingly the same hyperparameter setup produces entirely different results due to untracked experimental changes in the code, but not due to the variance of the algorithm.

\section{Algorithms}\label{sec:background}
To make \texttt{catalyst.RL} as effective as possible we closely follow the progress in the field of reinforcement learning. The following algorithms are available in our library (for brevity we only list the algorithms we used in the competition):
\begin{itemize}
    \item Deep Deterministic Policy Gradient (DDPG) \citep{lillicrap2015continuous}
    \item Twin Delayed Deep Deterministic Policy Gradient (TD3)~\citep{fujimoto2018addressing}
    \item  Soft Actor-Critic (SAC)~\citep{haarnoja2018soft}
    \item Distributional RL: categorical~\citep{bellemare2017distributional} and quantile~\citep{dabney2018distributional}
    \item Hyperbolic Discounting \citep{fedus2019hyperbolic}
\end{itemize}
Table~\ref{tab:frameworks} provides the result of comparison of \texttt{catalyst.RL} and other open--source frameworks. See Appendix~\ref{sec:app-rl} for a thorough discussion of these algorithms and Appendix~\ref{sec:app-related} for a discussion of various RL frameworks. 

We also provide default configuration files as a part of \texttt{catalyst.RL} so that users can easily run these algorithms on our included test environments. The modular structure of our framework allows one to easily combine various algorithms, so, for instance, an agent utilizing hyperbolic discounting along with quantile value approximation can be constructed by adding only a few lines to a config file. Another important feature is that the user can add his own exploration strategies, algorithms, and environments with minimal code required, and without changing the library core.

\begin{table}
\centering
\begin{tabular}{l|c|cccccc}
\toprule
& Catalyst.RL & OpenAI Baselines & Ray.rllib & Horizon &   Coach & SLM-Lab & Dopamine \\
\midrule
         Distributed &   \markc &           \markx &    \markc &  \markx &  \markc &  \markc &   \markx \\
                 DQN &   \markc &           \markc &    \markc &  \markc &  \markc &  \markc &   \markc \\
             Rainbow &   \markx &           \markx &    \markc &  \markx &  \markc &  \markx &   \markc \\
           REINFORCE &   \markc &           \markx &    \markc &  \markx &  \markx &  \markc &   \markx \\
                 A2C &   \markx &           \markc &    \markc &  \markx &  \markx &  \markc &   \markx \\
                 PPO &   \markc &           \markc &    \markc &  \markx &  \markc &  \markc &   \markx \\
                DDPG &   \markc &           \markc &    \markc &  \markc &  \markc &  \markx &   \markx \\
                 SAC &   \markc &           \markx &    \markx &  \markc &  \markc &  \markc &   \markx \\
                 TD3 &   \markc &           \markx &    \markc &  \markx &  \markc &  \markx &   \markx \\
              C51/QR &   \markc &           \markx &    \markc &  \markx &  \markc &  \markx &   \markc \\
 hyperbolic RL &   \markc &           \markx &    \markx &  \markx &  \markx &  \markx &   \markx \\
\bottomrule
\end{tabular}
\vspace{3pt}

\caption{Comparison of \texttt{catalyst.RL} with a number of RL frameworks. For a more detailed up-to-date table please check this Google Sheet\protect\footnotemark. }
\label{tab:frameworks}
\end{table}

\footnotetext{\url{https://docs.google.com/spreadsheets/d/1EeFPd-XIQ3mq_9snTlAZSsFY7Hbnmd7P5bbT8LPuMn0}.}

\section{Learning to Move Challenge}

% \begin{figure}[ht]
% \centering
% \includegraphics[width=0.48\textwidth]{img/final.png}
% \caption{AI for Prosthetics Challenge. Average reward of RL agent (TD3 with quantile value distribution approximation) trained with \texttt{catalyst.RL} (gray) and its rescaled version (blue) to better visualize the progress made after the first $20$ hours of training. Shaded region represents the standard deviation of the reward.}
% \label{fig:l2run}
% \end{figure}

We applied \texttt{Catalyst.RL} to the task of developing a controller to enable a physiologically-based human model to walk along a specified vector field.
This task was introduced at NeurIPS 2019 Learning to Move\footnote{\url{https://www.aicrowd.com/challenges/neurips-2019-learn-to-move-walk-around}}, where our team took the 2nd place, capitalizing on the sample-efficient off-policy algorithms implementations from \texttt{Catalyst.RL}. In this problem, the sample-efficiency was especially important as the simulator generating data was extremely slow ($\sim1$ observation per second).

\subsection{Task overview}

\begin{wrapfigure}{R}{0.5\textwidth}%
	\centering
	\includegraphics[width=0.49\textwidth]{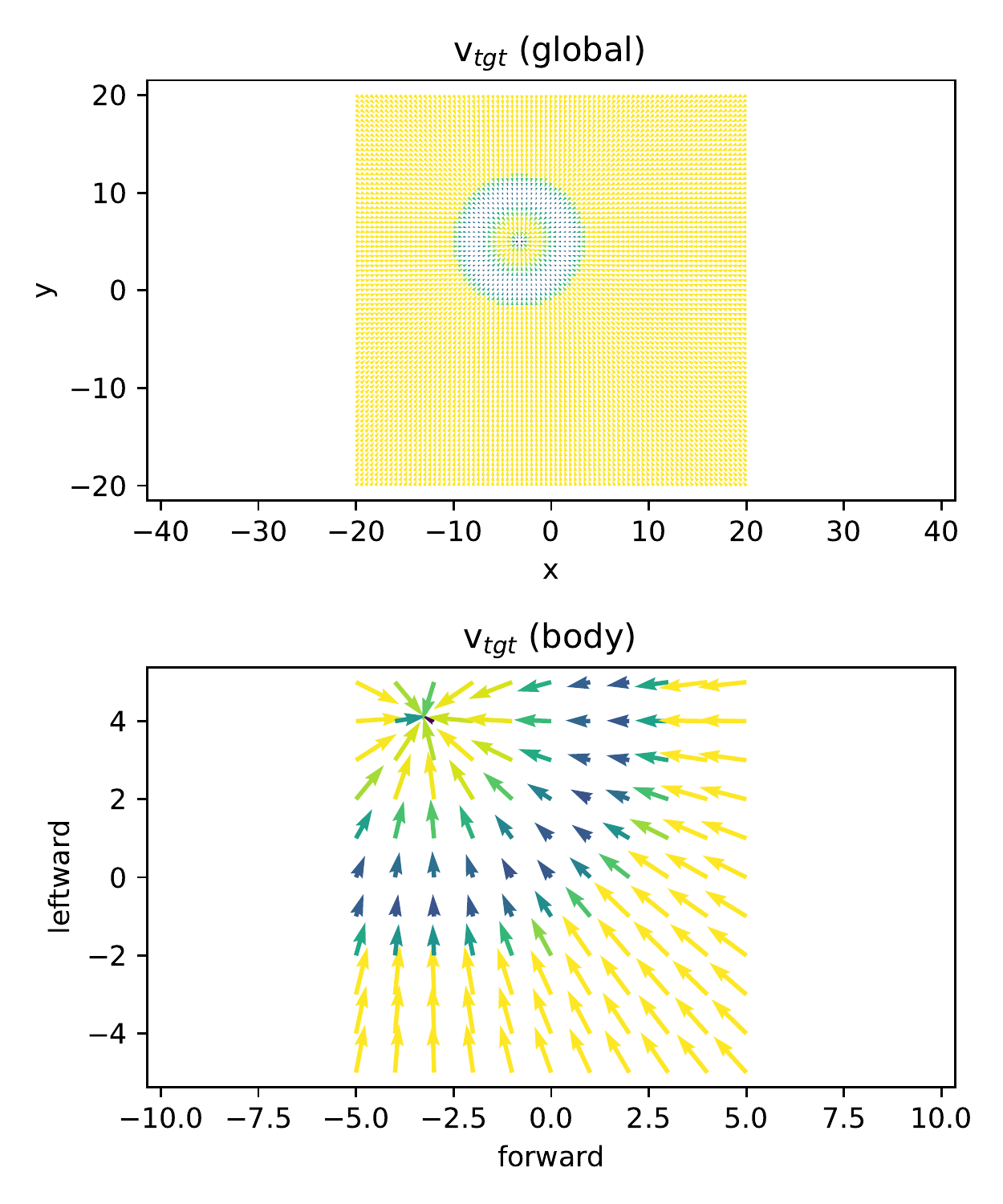}
	\caption{Example of a vector field in the environment. Global view (top) and local view (bottom).}
	\label{fig:vector_field}
	\vspace{-30pt}
\end{wrapfigure}%

The task was to build a controller which mapped states $\mathbf{s}_t \in \mathbb{R}^{97 + 2\times11\times11}$ to muscle activations $\mathbf{a}_t \in \mathbb{R}^{22}$. A state consisted of two parts: a 2D vector field $\mathbf{v}_t \in \mathbb{R}^{2 \times 11 \times 11}$ and remaining $97$ coordinates $\mathbf{p}_t$ representing positions and velocities of various muscles and pelvis. While the exact form of the reward is quite non-trivial, it was designed with the goal to force the agent to move along the given vector field, i.e., to match its velocity at a point $x$ with some fixed vector field evaluated at $x$. $\mathbf{v}_t$ represented samples from this vector field on a uniform grid around the agent (in the local frame of reference). Another term in reward was penalizing the agent for excessive muscle activations to enforce human-like gait.
The episode ends after $2500$ time steps or if the pelvis of the agent falls below $0.6$ meters. The target vector fields (in Round 2) had a \emph{sink} (see Figure~\ref{fig:vector_field}), i.e., there was a single attracting point of the field. Upon reaching this point the agent received a significant bonus reward after standing still for a short period of time, and in Round 2 a new target field appeared (totaling in two different vector fields the agent had to follow).

The ability of \texttt{catalyst.RL} to run multiple different algorithms and/or hyperparameter setups in parallel allowed us to quickly select the best candidate for the final solution -- the combination of TD3 algorithm with quantile distribution approximation and hyperbolic critic with $10$ value heads. See Appendix~\ref{sec:app-l2m} for specifics of reward shaping, state processing, and other implementation details.

\subsection{Approach overview}

\paragraph{Quantile TD3 with hyperbolic gammas}\label{sss:approach-solution}
As our baseline approach we use TD3 algorithm with 2 critics with quantile value function approximation on 101 atoms and hyperbolic gammas with 10 heads and $\gamma_{\max}=0.99$. To our knowledge, this is the first time such approach were used for continuous action control environments.

\paragraph{Hybrid exploration}\label{sss:approach-exploration}
We employ a hybrid exploration scheme which combines several heterogeneous types of exploration. 
% For each sampler episode we use only one specific exploration technique.
With $70\%$ probability, we add random noise from $\mathcal{N}(0,\sigma I)$ to the action produced by the actor where $\sigma$ changes linearly from $0$ to $0.3$ for different sampler instances. With $20\%$ probability we apply parameter space noise~\citep{plappert2017parameter} with adaptive noise scaling, and we do not use any exploration otherwise. The decision of which exploration scheme to choose is made at the beginning of the trajectory and is not changed till its end.

\paragraph{Ensemble learning}\label{sss:approach-ensemble}
Additional important technique of our solution was agents ensemble learning with shared replay buffer. We take our ``baseline'' approach and vary its hyperparameters to get several ``replicas''. Then we simultaneously start agents learning with shared trajectories database – by this way, they effectively share all gained experience. In our experiments we vary ``history length'', number of previous observation used for decision making, hyperparameter – from 1 to 12.

% \textcolor{red}{
% \paragraph{Action postprocessing}
% In order to further minimize the muscle activations of the trained agent, we have used the following simple action postprocessing.
% $$
% \mathbf{a}^{\mathrm{clipped}}_t = \begin{cases} \mathrm{clip}(\mathbf{a}_t, 0, 0.7), \quad \|\mathbf{v}^{\mathrm{cur}_t \|} > 0.3 \\
% \mathrm{clip}(\mathbf{a}_t, 0, 0.3), \quad \|\mathbf{v}^{\mathrm{cur}_t \|} \leq 0.3 \\
% \end{cases}
% $$
% }

\subsection{Experiments and results}
\begin{figure}[htb!]
     \centering
     \includegraphics[width=0.9\textwidth]{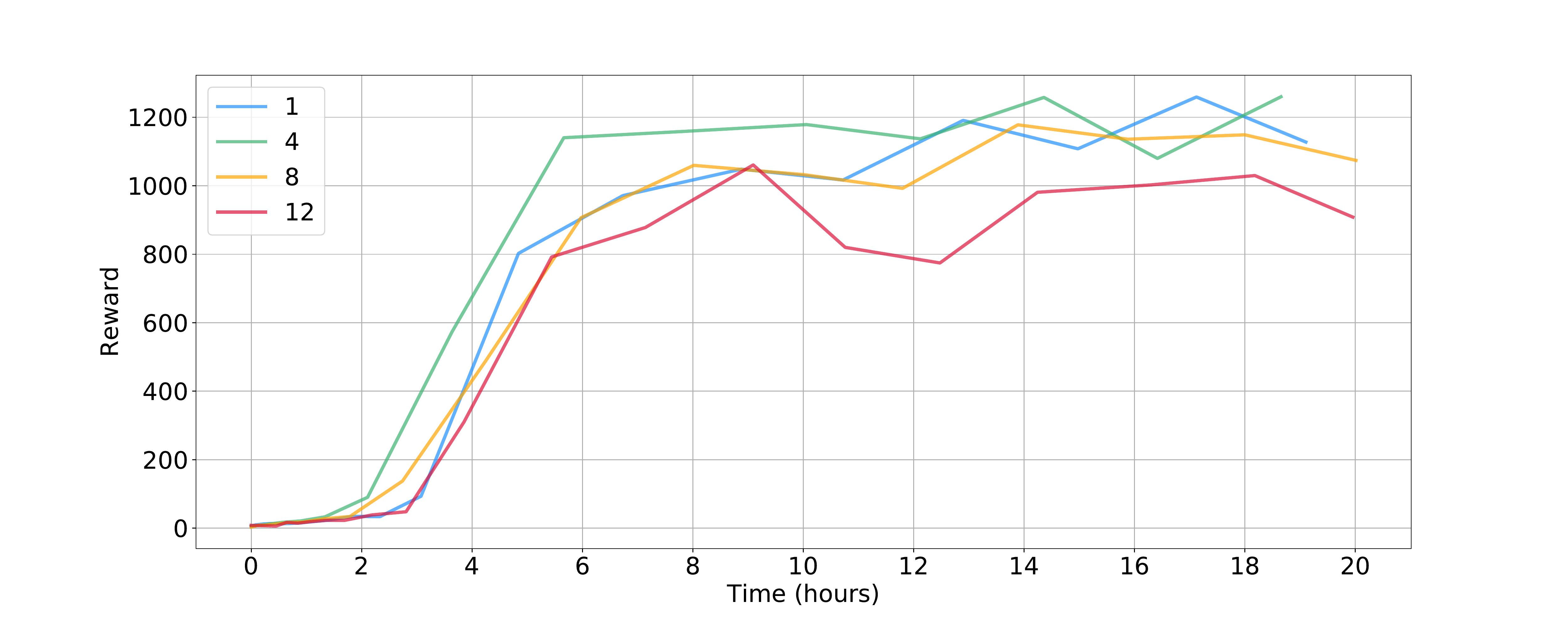}
     \caption{Average raw reward on 64 validation seeds versus training time (in hours). Different graphs correspond to different history lengths used for training.}
     \label{fig:rewards}
 \end{figure}

During our experiments, we have evaluated training performance of different models. For the complete list of hyperparameter and their values we refer the reader to our GitHub repository. Results of individual agents of our best performing  ensemble are presented on Figure~\ref{fig:rewards}. We have used TD3 with quantile value approximation and hyperbolic critics with 10 heads and $\gamma_{\max}=0.99$. Four presented agents were trained with frame skip being $4$ and history length $1,4,8,12$ correspondingly. Our ensemble learning technique exhibits remarkable sample efficiency exceeding a score of $1000$ after just $6$ hours of training with $40$ parallel CPU threads ($24$ for samplers and $16$ for trainers).

\section{Conclusion}\label{sec:conclusion}
In this paper, we introduced \texttt{catalyst.RL}, an open-source library for RL research, which offers a variety of tools for the development and evaluation of RL algorithms.

Our results in OpenSim experiments indicate that in computationally expensive stochastic environments that have high-dimensional continuous action space the best performing method are off-policy DDPG-based approaches. We have tested several modifications to TD3 and each turned out to be important for learning. 
Quantile value function approximation allows us more precise state value prediction. Hyperbolic gammas give us the ability to learn several multi-horizon sub-policies simultaneously. Parameter and Layer noise additionally improves the stability of learning due to the introduction of state-dependent exploration. 
Finally, ensemble learning
% with shared replay buffer 
over contrasting agents with different hyperparameters additionally improves the exploration of the environment.
In this way, the agent truly learns to run and walk in only a few hours of training time, which the exhibits remarkable sample efficiency of the proposed approach.

Examples of the learned policies are present at this URL \url{https://youtu.be/WuqNdNBVzzI}. Source code is available at \url{https://github.com/Scitator/run-skeleton-run-in-3d}.
In general, we believe that the investigation of human-based agents in physically realistic environments is a promising direction for future research.

\paragraph{Future work} In the future, we plan to continue the development and support of \texttt{catalyst.RL}, as well as to heavily benchmark all proposed approaches and get a better understanding of the applicability of certain methods in deep reinforcement learning.

%\clearpage
%% The file named.bst is a bibliography style file for BibTeX 0.99c
\bibliographystyle{named}
\bibliography{ijcai19}
\newpage
\appendix
\section{RL background}{\label{sec:app-rl}}
In this section, we introduce necessary notation and briefly revise RL algorithms for continuous control we use in our experiments.

\paragraph{RL problem statement} In reinforcement learning, the agent interacts with the environment modeled as a Markov Decision Process (MDP). MDP is a five-tuple $\left(\mathcal{S}, \mathcal{A}, r, P, \gamma \right)$ where $\mathcal{S}$ and $\mathcal{A}$ are the state and action spaces, $r:\mathcal{S}\times\mathcal{A}\rightarrow\mathbb{R}$ is the reward function, $P:\mathcal{S}\times\mathcal{A}\times\mathcal{S}\rightarrow\mathbb{R}_{\geq 0}$ denotes probability density of transitioning to the next state $\mathbf{s}_{t+1} \in \mathcal{S}$ from the current state $\mathbf{s}_{t} \in \mathcal{S}$ after executing action $\mathbf{a}_t \in \mathcal{A}$, and $\gamma \in [0,1]$ is the discount factor. The goal of RL is to maximize the expected return.

In our formulation the action space is continuous $\mathcal{A}\subseteq\mathbb{R}^d$ and the agent decides what action to take according to either deterministic $\mu_\theta:\mathcal{S}\rightarrow\mathcal{A}$ or stochastic $\pi_\theta:\mathcal{S}\rightarrow\left(\mathcal{A}\rightarrow\mathbb{R}_{\geq 0}\right)$ policy represented as neural net with parameters $\theta$. $Z_\phi(\mathbf{s}_t,\mathbf{a}_t) \approx \sum_{t'>=t} \gamma^{t'-t}r(\mathbf{s}_{t'},\mathbf{a}_{t'})$ stands for the approximation of discounted return and $Q_\phi(\mathbf{s}_t,\mathbf{a}_t) = \mathbb{E}Z_\phi(\mathbf{s}_t,\mathbf{a}_t)$ stands for Q-function approximation.

\paragraph{DDPG} Deep Deterministic Policy Gradient (DDPG) is off-policy reinforcement learning algorithm applicable to continuous action spaces~\citep{lillicrap2015continuous}. It usually consists of two neural nets: one of them (actor) approximates deterministic policy and the other (critic) is used for Q-function approximation. The parameters of these networks are updated by the following rules:
\begin{equation*}
\begin{aligned}
\phi \leftarrow \phi &- \alpha \nabla_{\phi} Q_\phi(\mathbf{s}_t,\mathbf{a}_t) \left( y^\text{DDPG} - Q_{\phi}(\mathbf{s}_t,\mathbf{a}_t) \right), \\
\theta \leftarrow \theta &+ \beta \nabla_{\mathbf{a}_t} Q_\phi(\mathbf{s}_t,\mathbf{a}_t) \nabla_{\theta} \mu_{\theta}(\mathbf{s}_t),
\end{aligned}
\end{equation*}
where $y^\text{DDPG} = r_t + \gamma Q_{\tilde\phi}(\mathbf{s}_{t+1},\mu_\theta(\mathbf{s}_{t+1}))$ are TD-targets and transitions $(\mathbf{s}_t,\mathbf{a}_t,r_t,\mathbf{s}_{t+1})$ are uniformly sampled from the experience replay buffer $\mathcal{D}$~\citep{mnih2015human}. TD-targets make use of a target network $Q_{\tilde\phi}$, the architecture of which copies the critic's architecture, and its weights are updated by slowly tracking the learned critic: 
\[\tilde\phi \leftarrow \tau\phi + (1-\tau)\tilde\phi, \quad \tau \ll 1.
\]

\paragraph{TD3} Twin Delayed Deep Deterministic Policy Gradient (TD3)~\citep{fujimoto2018addressing} is a recent improvement over DDPG which adopts Double Q-learning technique to alleviate overestimation bias in actor-critic methods. The differences between TD3 and DDPG are threefold. Firstly, TD3 uses a pair of critics which provides pessimistic estimates of Q-values in TD-targets
\[
Q(\mathbf{s}_{t+1},\mathbf{a}_{t+1}) = \min_{i=1,2} Q_{\tilde{\phi}_i}(\mathbf{s}_{t+1},\mathbf{a}_{t+1}).
\]
Secondly, TD3 introduces a novel regularization strategy, target policy smoothing, which proposes to fit the value of a small area around the target action
\begin{equation*}
\begin{aligned}
y^\text{TD3}&= r_t + \gamma Q(\mathbf{s}_{t+1},\mathbf{a}_{t+1}), \\
\mathbf{a}_{t+1} &= \mu_\theta(\mathbf{s}_{t+1}) + \boldsymbol{\epsilon},\quad\boldsymbol{\epsilon}\sim \text{clip}(\mathcal{N}(0,\sigma I),-c,c).
\end{aligned}
\end{equation*}
Thirdly, TD3 updates an actor network less frequently than a critic network (typically, one actor update for two critic updates). 

In our experiments, the application of the first two modifications led to much more stable and robust learning. 
Updating the actor less often did not result in better performance, thus, that modification was omitted in the experiments. However, our implementation allows to include it back.

\paragraph{SAC} Soft Actor-Critic (SAC)~\citep{haarnoja2018soft} optimizes more general maximum entropy objective and learns a stochastic policy $\pi_\theta(\mathbf{a}_t|\mathbf{s}_t)$, usually parametrized in a way to support reparametrization trick (e.g., Gaussian policy with learnable mean and variance or policy represented via normalizing flow) and a pair of Q-functions. The parameters of all networks are updated by the following rules:

\begin{equation*}
\begin{aligned}
\phi_i &\leftarrow \phi_i - \alpha \nabla_{\phi_i} Q_{\phi_i}(\mathbf{s}_t,\mathbf{a}_t) \left( y^\text{SAC} - Q_{\phi_i}(\mathbf{s}_t,\mathbf{a}_t) \right), \\
\theta &\leftarrow \theta + \beta \nabla_{\theta} \left( Q_{\phi_1}(\mathbf{s}_t,\mathbf{a}_t) - \log\pi_\theta(\mathbf{a}_t|\mathbf{s}_t) \right), \\
y^\text{SAC} &= r_t + \gamma ( \min_{i=1,2} Q_{\tilde{\phi}_i}(\mathbf{s}_{t+1},\mathbf{a}_{t+1}) -\log\pi_\theta(\mathbf{a}_{t+1}|\mathbf{s}_{t+1})).
\end{aligned}
\end{equation*}

\paragraph{Distributional RL} A distributional perspective on RL suggests to approximate the distribution of discounted return instead of the approximation its first moment (Q-function) only. In this work, we analyze two different ways to parametrize this distribution (also referred to as \textit{value distribution}), namely categorical~\citep{bellemare2017distributional} and quantile~\citep{dabney2018distributional}.

Categorical return approximation represents discounted return as $Z_\phi(\mathbf{s}_t,\mathbf{a}_t) = \sum_{i=1}^N p_{\phi,i}(\mathbf{s}_t,\mathbf{a}_t) \delta_{z_i}$, where $\delta_z$ denotes a Dirac at $z \in \mathbb{R}$ and atoms $z_i$ split the distribution support $[V_\text{MIN},V_\text{MAX}] \subset \mathbb{R}$ by $N-1$ equal parts. Its parameters are updated by minimizing the cross-entropy between $Z_\phi(\mathbf{s}_t,\mathbf{a}_t)$ and $\Phi\tilde{\mathcal{T}} Z_{\tilde{\phi}}(\mathbf{s}_t,\mathbf{a}_t)$ --- the Cramer projection of target value distribution $\tilde{\mathcal{T}} Z_{\tilde{\phi}}(\mathbf{s}_t,\mathbf{a}_t) = r_t + \gamma Z_{\tilde{\phi}}(\mathbf{s}_{t+1},\mathbf{a}_{t+1})$ onto the distribution support.

Quantile approximation sticks to the ``transposed'' parametrization $$Z_\phi(\mathbf{s}_t,\mathbf{a}_t) = \frac{1}{N}\sum_{i=1}^N \delta_{z_{\phi, i}(\mathbf{s}_t,\mathbf{a}_t)},$$ which assigns equal probability masses $\frac{1}{N}$ to the set of learnable atom positions $\{z_{\phi,i}(\mathbf{s}_t,\mathbf{a}_t)\}_{i=1}^N$. Distribution parameters are updated by minimizing the quantile regression Huber loss. The exact formulas are cumbersome and omitted here for brevity, thus we refer the interested reader to~\citep{dabney2018distributional}.

\paragraph{Hyperbolic Discounting}
Typically, RL algorithms include a discount factor $\gamma$ which leads to an exponentially decaying values of future rewards. However, evidence from human psychology and economics suggests that humans instead have hyperbolic time-preferences: the value of the reward at time $T$ decays by the rule $\frac{1}{1 + kT}$ rather than exponentially.
In a recent work \citep{fedus2019hyperbolic} authors propose a new approach to incorporate this observation into standard temporal-difference learning techniques by learning value functions over multiple time-horizons. We refer the reader to the original paper for derivations and simply discuss details necessary for the implementation.
We start with the \emph{hyperbolic discount function}
$$\Gamma(t) = \frac{1}{1 + kt}$$ which can be conveniently expressed as weighting over exponential discount functions $\gamma^t$:
\begin{equation}{\label{eq:gamma}}
 \Gamma(t) = \int_{\gamma=0}^1 w(\gamma) \gamma^t d \gamma,   
\end{equation}
with weights $w(\gamma) = \frac{1}{k}\gamma^{\frac{1}{k}-1}$. The corresponding \emph{hyperbolic Q-function} can be written as a weighting over standard exponentially-discounted Q-functions using the same weights $w(\gamma)$.

$$
Q^{\Gamma}_{\pi}(s, a) = \int_{\gamma=0}^1 w(\gamma) Q^{\gamma}_{\pi}(s, a) d\gamma
$$

The integral in \eqref{eq:gamma} can be approximated with a Riemann sum over the discrete intervals
$$
\mathcal{G} = [\gamma_0, \gamma_1, \ldots, \gamma_N],
$$
namely
$$
Q^{\Gamma}_{\pi}(s, a) \approx \sum_{\gamma_i \in \mathcal{G}} (\gamma_{i+1} - \gamma_{i}) w_{\gamma_i} Q^{\gamma_i}_{\pi}(s, a).
$$
This formula suggests that to learn the hyperbolic Q-function we can simultaneously learn standard Q-functions but for multiple values of discount factor $\gamma$. In practice, we can implement this approach by setting the critic to have multiple output \emph{heads}, with each head to corresponding to a particular value of gamma, and apply standard learning algorithms.
This multi-horizon model predicts Q-values for $n$ separate discount functions thereby modeling different effective horizons. Each Q-value is a lightweight computation, an affine transformation of a shared representation. By modeling over multiple time-horizons, we now have the option to construct policies that act according to a particular value or a weighted combination. Note that while this theory is less justified for typical continuous control settings, where a more rigorous approach would also include multi-head \emph{actors} (predicting the optimal actions for each time-horizon), we found that even such a modifications improved performance of our algorithms. Following \citep{fedus2019hyperbolic} for training we took gammas forming a log-uniform grid on the interval $[0, \gamma_{max}]$, and for inference, we only used the head corresponding to the largest gamma.

\section{Learning to Move}\label{sec:app-l2m}
\subsection{Reward shaping and processing}
\paragraph{Sparse reward}
Reward in the environment was designed in such a way that the agent received it only on successful \emph{footsteps}, which are presumably quite rare events. To alleviate the problem with reward sparsity, we heavily used reward shaping and introduced various types of \emph{dense} rewards for the agent.
\paragraph{Vector field reward shaping} We start with various modifications designed to provide the agent with better signal relating its velocity and the target vector field. 
Let $\mathbf{v}^{\mathrm{cur}}_t$ denote \mbox{$\mathbf{v}_t[:,5,5]$}, i.e., the target vector field evaluated at the position of the agent. Then extra reward in the form
$$
\mathbf{r}^{\mathrm{vec}}_{t} = \max \lbrace 1 - A^2_{\mathrm{}} \|\mathbf{v}^{\text{pelvis}} - \mathbf{v}_t^{\mathrm{cur}}\|^2, 0 \rbrace, 
$$
directly motivates motion along the target vector field (where $A$ chosen to be the reciprocal of the maximal possible difference in velocities).

Another approach that we tried is to reward the correct direction and magnitude separately.
$$
\mathbf{r}^{\mathrm{vel}}_{t} = \max \lbrace 1 - B^2 |\|\mathbf{v}^{\text{pelvis}}\| - \|\mathbf{v}_t^{\mathrm{cur}}\| |^2, 0 \rbrace,
$$
$$
\mathbf{r}^{\mathrm{dir}}_{t} = \max \lbrace 1 - C^2 \Big\lVert \frac{\mathbf{v}^{\text{pelvis}}}{\| \mathbf{v}^{\text{pelvis}} \|} - \frac{\mathbf{v}_t^{\mathrm{cur}}}{\| \mathbf{v}_t^{\mathrm{cur}} \| }\Big\rVert^2, 0 \rbrace.
$$
Thus, the total velocity bonus reward we used in our solution had the following form.
$$
\mathbf{r}^{\mathrm{velocity}}_{t} = R_1 \mathbf{r}^{\mathrm{vec}}_{t} + R_2 \mathbf{r}^{\mathrm{vel}}_{t} + R_3 \mathbf{r}^{\mathrm{dir}}_{t}
$$
\paragraph{Target reward shaping}
A large part of total reward comes from the ability of the agent to stand still for several seconds within a vicinity of the sink of the target vector field. While in principle the agent trained to follow the vector field sufficiently well should obtain it automatically, we found that providing the agent with extra reward for minimizing distance to the sink makes it much easier. More specifically, we used the reward bonus of the following form:
$$
\mathbf{r}^{\mathrm{target}}_{t} = 1 - D^2 \|\mathbf{p}^{\mathrm{body}} - \mathbf{p}^{\mathrm{sink}}\|^2.
$$
\paragraph{Biomechanical reward shaping}
At the early stages of the competition we have noticed that sometimes the learned agent tended to cross its
legs as the simulator allowed one leg to pass through another. We have assumed that such behavior led to
suboptimal policy and excluded it by introducing additional ``crossing legs'' penalty. Specifically, we have
computed the scalar triple product of three vectors starting at pelvis and ending at head, left toe, and right
toe, respectively, which resulted in a penalty of the following form.
$$
\mathbf{r}^{\mathrm{crossing\_legs}}_{t} = -E^2 \min \lbrace (\mathbf{r}^{\mathrm{head}} - \mathbf{r}^{\mathrm{pelvis}}, \mathbf{r}^{\mathrm{left}} - \mathbf{r}^{\mathrm{pelvis}}, 
\mathbf{r}^{\mathrm{right}} - \mathbf{r}^{\mathrm{pelvis}}), 0 \rbrace
$$
In order to encourage the agent to learn human-like gaits, we have also used the knee bending bonus, providing the agent with an extra reward proportional to knee flexion.
\paragraph{State postprocessing}
While in principle $\mathbf{v}_t$ contains useful information about the target vectors in a neighbourhood of the agent, we found that using simply the central vector $\mathbf{v}^{\mathrm{cur}}$ is sufficient to learn good policies. Thus, to obtain the observation we concatenated $\mathbf{v}^{\mathrm{cur}}_t$ and $\mathbf{p}_t$. We have also rescaled the time step index into a real number in $[-1, 1]$ and included it into the observation vector, resulting in the observation of size $100$.

\subsection{Additional improvements}
In order to gain the best performance from several learned actor-critic pairs, we employ a number of stabilizing improvements.
\paragraph{Checkpoints ensemble} Adapting the ideas from \citep{huang2017learning,dietterich2000ensemble,huang2017snapshot, DBLP:journals/corr/abs-1902-02441} and capitalizing on our distributed framework, we simultaneously train several instances of our algorithm with different sets of hyperparameters, and
then pick the best checkpoints in accordance to validation runs on 80 random seeds. Given an ensemble of
actors and critics, each actor proposes the action which is then evaluated by all critics. After that, the action
with the highest average Q-value is chosen.
\paragraph{Action mixtures} In order to extend our action search space, we also evaluate various linear combinations
of the actions proposed by the actors. This trick slightly improves the resulting performance at no additional
cost as all extra actions are evaluated together in a single forward pass.

\section{Frameworks for Reinforcement Learning}\label{sec:app-related}
To date, more than twenty reinforcement-learning-themed open-source software libraries have been
released. \texttt{catalyst.RL} is most similar to libraries that implement a wide variety of algorithms and are intended to be applied to a variety of RL problems. Some of the following libraries are: OpenAI Baselines \citep{baselines} with reference implementations of various RL algorithms, Ray.rllib \citep{liang2017ray}, an open-source library for RL that offered both a collection of reference algorithms and scalable primitives for composing new ones, Horizon \citep{gauci2018horizon} Facebook's open-source applied RL platform, Coach \citep{caspi_itai_2017_1134899}, SLM-Lab \citep{anonymous2020slm}, Dopamine \citep{castro2018dopamine}, a research framework for fast prototyping of reinforcement learning algorithms in \texttt{TensorFlow}. 

\end{document}